\documentclass[sigconf]{acmart}

\AtBeginDocument{%
  \providecommand\BibTeX{{%
    \normalfont B\kern-0.5em{\scshape i\kern-0.25em b}\kern-0.8em\TeX}}}

\acmYear{2023}
\setcopyright{acmlicensed}\acmConference[WWW '23]{Proceedings of the ACM Web Conference 2023}{May 1--5, 2023}{Austin, TX, USA}
\acmBooktitle{Proceedings of the ACM Web Conference 2023 (WWW '23), May 1--5, 2023, Austin, TX, USA}
\acmPrice{15.00}
\acmDOI{10.1145/3543507.3583201}
\acmISBN{978-1-4503-9416-1/23/04}

\usepackage{dsfont}

\usepackage{amsmath,amssymb}
\usepackage{algpseudocode,algorithm}
\algtext*{EndFor}%

\newcommand{\mymethod}{EvCBR} 

\begin{document}

\title{Event Prediction using Case-Based Reasoning over Knowledge~Graphs}

\author{Sola Shirai}
\authornote{Part of this work was done while at IBM Research.}
\orcid{0000-0001-6913-3598}
\affiliation{%
  \institution{Rensselaer Polytechnic Institute}
  \city{Troy}
  \state{NY}
  \country{USA}
  \postcode{12180}
}
\email{shiras2@rpi.edu}

\author{Debarun Bhattacharjya}
\orcid{0000-0002-9125-1336}
\affiliation{%
  \institution{IBM Research}
  \city{Yorktown Heights}
  \state{NY}
  \country{USA}
}
\email{debarunb@us.ibm.com}

\author{Oktie Hassanzadeh}
\orcid{0000-0001-5307-9857}
\affiliation{%
  \institution{IBM Research}
  \city{Yorktown Heights}
  \state{NY}
  \country{USA}
}
\email{hassanzadeh@us.ibm.com}


\begin{abstract}
Applying link prediction (LP) methods over knowledge graphs (KG) for tasks such as causal event prediction presents an exciting opportunity. 
However, typical LP models are ill-suited for this task as they are incapable of performing inductive link prediction for new, unseen event entities and they require retraining as knowledge is added or changed in the underlying KG. 
We introduce a case-based reasoning model, \mymethod{}, to predict properties about new consequent events based on similar cause-effect events present in the KG. \mymethod{} uses statistical measures to identify similar events and performs path-based predictions, requiring no training step.
To generalize our methods beyond the domain of event prediction, we frame our task as a 2-hop LP task, where the first hop is a causal relation connecting a cause event to a new effect event and the second hop is a property about the new event which we wish to predict.
The effectiveness of our method is demonstrated using a novel dataset of newsworthy events with causal relations curated from Wikidata, where \mymethod{} outperforms baselines including translational-distance-based, GNN-based, and rule-based LP models.
\end{abstract}



\maketitle

\section{Introduction}

In recent years, the use of knowledge graphs (KG) in various applications has become increasingly common. The coverage and scope of domain-specific KGs continues to grow, and more general-purpose KGs such as Wikidata are capable of representing enormous amounts of knowledge.
While a variety of methods exist that can leverage KGs to perform tasks, one of the most prevalent methods is to perform KG completion -- and more specifically, link prediction (LP) -- to glean new knowledge. Since we can expect KGs to not have \textit{complete} coverage of all available knowledge \cite{Paulheim2017KnowledgeGR}, LP is a valuable tool to apply and refine the contents of a KG.

\begin{figure}
    \centering
    \includegraphics[width=\linewidth]{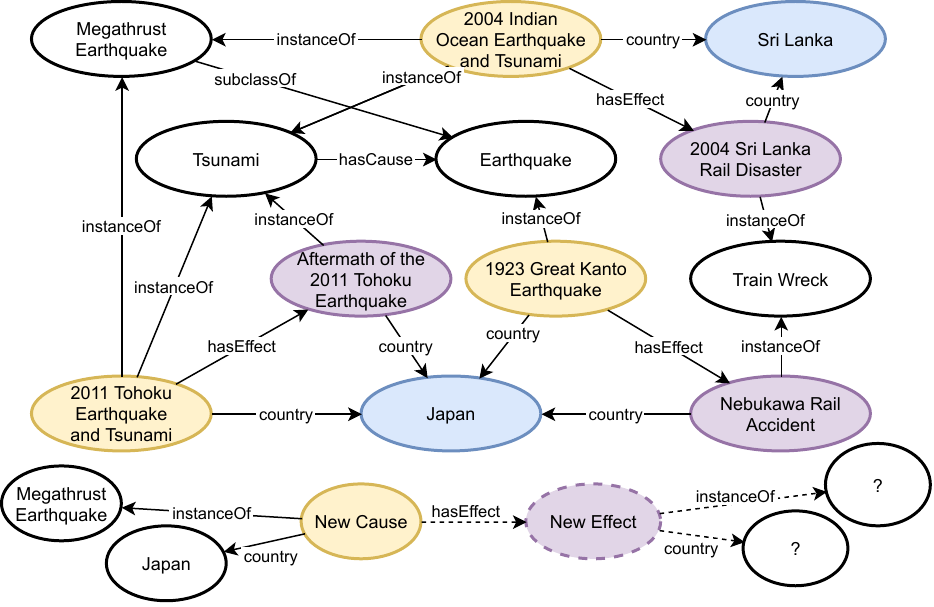}
    \caption{A running example depicting a snippet of Wikidata containing causal events as well as a new cause event and effect event for which we wish to make predictions.}
    \label{fig:motivating_example}
\end{figure}

In this work we aim to apply a KG-driven approach to the task of event prediction for news events as a motivating example. Consider Figure \ref{fig:motivating_example}, which depicts a snippet of Wikidata \cite{Wikidata} surrounding several events. In addition to properties about events such as their \texttt{instanceOf} and \texttt{country}, we can also observe a number of causal relations between events. For example, the Kanto Earthquake is connected by the \texttt{hasEffect}\footnote{\url{https://www.wikidata.org/wiki/Help:Modeling_causes}} relation to the Nebukawa Rail Accident, indicating that the earthquake caused the train wreck event.

As a running example, suppose we wish to predict the effect of a new Megathrust Earthquake event that occurred in Japan. We might represent this earthquake as a new entity, New Cause, which has a causal relation to another new entity, New Effect, as depicted in Figure \ref{fig:motivating_example}. In order to predict properties about New Effect, such as its \texttt{instanceOf} and \texttt{country}, we can apply LP methods to connect New Effect to the appropriate entities in the KG. 

However, a major limitation of most LP methods is that relations can only be predicted among entities that already exist in the KG. A more practical and challenging perspective for this task is to consider LP for entirely new entities (i.e., inductive link prediction). This perspective also better reflects the needs of an event prediction system, which would aim to make predictions about entirely new events. The inductive setting introduces an additional level of difficulty for LP models, as many of them (especially embedding-based ones) operate under the closed world assumption -- that is, they can only perform predictions for entities that have been seen during training, and facts that are not seen during training are assumed to be false.
Additionally, even for models that are capable of this task, some KGs (such as Wikidata) constantly undergo updates and changes, which can negatively affect a model's performance as its training data becomes outdated.

Towards our goal of applying KGs to perform event prediction, we develop a case-based reasoning approach, \mymethod{},
which leverages examples of past events in the KG to perform LP for the properties of unseen event entities.
We frame our problem as a 2-hop link prediction task -- e.g., starting from a cause event, we assume the existence of a causal relation to a new effect event and predict its properties. 
This approach allows us to perform inductive link prediction for the properties of the unseen effect without the need for external data. 
Importantly, \mymethod{} requires no training, relying instead on statistical measures and subclass hierarchies to identify similar entities and follow paths through the KG to perform predictions. This makes \mymethod{} well-suited to KGs such as Wikidata that frequently experience changes in their structure and content. More generally, we can also apply \mymethod{} to perform 2-hop LP for properties of any type of unseen entity that is connected to a known entity by some relation. 


Our contributions are as follows:
\begin{enumerate}
    \item We introduce a case-based reasoning model, \mymethod{},\footnote{We publicly release our code at \url{https://github.com/solashirai/WWW-EvCBR}} for event prediction. Our model considers this task as a LP task in a 2-hop setting, leveraging knowledge about similar cause-effect events to make predictions about the unseen effect. \mymethod{} requires no training, making it well-suited to handle new events and changes to the underlying KG.
    \item Compared to similar work, we introduce novel similarity metrics to identify similar cause-effect event cases as well as a 
    refinement step to improve the precision of predictions.
    \item We curate and release a novel dataset surrounding causal events in Wikidata, extracting news events that are connected by causal relations as well as their local connections.\footnote{Dataset available at \url{https://doi.org/10.5281/zenodo.7196049}.}
    \item In our 2-hop inductive link prediction task, our model shows superior performance on our event dataset as well as competitive performance on a modified evaluation dataset based on the FB15k-237 dataset.
\end{enumerate}

\section{Related Work}

Zhao~\cite{zhao2021event} presents a comprehensive survey of different kinds of event prediction methods across different domains. Under their taxonomy of event prediction methods~\cite[Fig. 3]{zhao2021event}, our method is most closely related to the ``causality-based" methods under ``semantic prediction" techniques. Other notable work that falls under this class of methods includes Radinsky et al's Pundit algorithm~\cite{DBLP:conf/www/RadinskyDM12} which is based on automated construction of a causality graph through extraction of causal patterns and generalization. Zhao et al.~\cite{zhaoCausalityNetwork} also construct a causal graph from text documents and build embeddings to perform a simple (one-hop) prediction. While prior work has explored applying LP methods to KGs of causal events \cite{Khatiwada2022,Shirai2022RuleBasedLP}, to our knowledge, our work is the first to apply 2-hop LP over a knowledge graph for event prediction. 

In the space of LP in KGs, a significant number of embedding-based methods have been developed as detailed in recent surveys \cite{wang2017knowledge,wang21survey}. Such methods often show trade-offs in performance based on the dataset 
and how well the model hyperparameters are tuned \cite{Teach2020YouCT}. While GNN-based models \cite{Zhang2022GNN} have often shown recent state-of-the-art performance, there has been discussion surrounding what semantics are actually captured in such models \cite{Jain2021DoEA} as well as re-evaluations of \textit{how} one evaluates such LP models \cite{tiwari21revisiting,Akrami2020RealisticRO,Sun2020ARO}. Furthermore, most embedding-based models are trained and evaluated in the closed world setting, with a relatively sparse number of recent works such as \cite{daza21inductive,ali21hyper} considering inductive link prediction using additional knowledge sources such as text or hyper-relational facts.

While less prevalent than embedding-based models, there are also a variety of rule-based LP models \cite{galarraga13AMIE,AMIE3,RuleN,RUDIK,DRUM}.
Methods such as AnyBURL \cite{AnyBURL} have shown similar performance to state-of-the-art embedding models while requiring significantly less training time \cite{rossi2021comparative}.
Works such as \cite{ProbCBR,akbcCBR} present case-based reasoning models which, like our method, require no training and make path-based predictions. Our method differs from such similar works in our method of computing entity similarity, approach to apply and refine prediction paths, and task formulation of performing 2-hop LP.


\section{Problem formulation}

Our work considers the application of LP methods over a knowledge graph for the task of event prediction. We define a knowledge graph $G=(E,R,T)$ as consisting of a set of entities $E$, a set of relations $R$, and a set of triples $T$. $T$ consists of triples of the form ($h$, $r$, $t$), where $h,t \in E$ and $r \in R$.

The task of LP is then to correctly predict the missing tail entity given a partial triple ($h$, $r$, $?t$).\footnote{We note the related LP tasks of predicting a missing head $?h$ or relation $?r$, but in this paper we focus on LP for predicting $?t$ as it aligns best with our event prediction task.} The standard assumption in this task is that the head entity $h$ and the missing tail entity $?t$ are known entities in $E$. To perform inductive link prediction, it is necessary to predict the missing $?t$ where $?t \notin E$. Since LP models 
can not predict
such an entity $?t$ solely from the content of $G$, inductive link prediction methods often also rely on additional background data (such as textual descriptions) to extract and represent new entities.

\subsection{Event Prediction Task
}

Extending the task of LP to event prediction, our goal is to make predictions about new events based on a causal relation between two events. We represent our task as a query triple $(c, r, e)$, where $c$, $r$, and $e$ denote the cause event, causal relation, and effect event, respectively. 
Furthermore, the point of the event prediction task is to predict properties about the new effect event -- i.e., predicting outgoing properties about $e$ in the form of $(e, r_e, ?z) \in Out_e$, where $r_{e} \in R$ and $?z \in E$. 
In this task, we assume that $c \in E$ while $e \notin E$ -- if we wish to make predictions about a cause $c$ which was not in the original KG, we assume that triples indicating the properties of $c$ are supplied as input and added to the KG. Additionally, we assume that we are given the relations $r_e$ to predict.

\paragraph{Example.}
In Figure \ref{fig:motivating_example}, our task is to perform predictions for New Effect based on the triple (New Cause, \texttt{hasEffect}, New Effect). Further, we make predictions about each of New Effect's properties, as (New Effect, \texttt{instanceOf}, $?z$) and (New Effect, \texttt{country}, $?z$).
\qed

To make predictions under these assumptions, we must consider two points: first, we must make a prediction for an unseen effect event $e$ given a cause $c$ of interest and a relation $r$; second, from the unseen event $e$, we must make predictions about its properties by performing LP to tail entities $?z \in E$.

Given this problem formulation for event prediction, we can then perform predictions for properties of $e$ without the need for any background knowledge about $e$ by performing a 2-hop LP task starting from $c$ -- that is, predicting the tail entity for the 2-hop link ($c$, $r$, $r_e$, $?z$) for each triple in $Out_e$. By omitting explicit consideration of $e$ in the prediction task, we are able to overcome the limitation that LP models must have a way of extracting or representing the entity. From a practical perspective, when performing event predictions, one need not explicitly represent $e$ as the primary focus is on predicting properties about the effect event.

\begin{figure}[htbp]
    \centering
    \includegraphics[width=\linewidth]{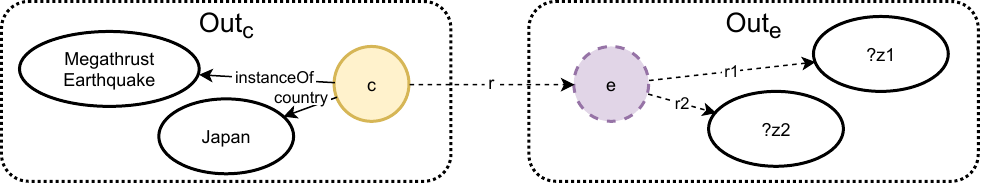}
    \caption{The task setup for our running example.}
    \label{fig:runningExSetup}
\end{figure}

We additionally denote the set of outgoing triples from $c$ as $Out_c$ -- we assume that outgoing properties of the cause event $c$ are provided or already present in the KG, and our method only makes use of such outgoing triples\footnote{We are motivated to only consider outgoing edges because we can expect a new event entered into the KG to have little to no incoming relations present.} from $c$. A graphical depiction of this problem setup for our running example is shown in Figure \ref{fig:runningExSetup}.

\subsection{A General 2-Hop Prediction Task}

We pose a generalization of the event prediction task, which could be useful in other applications: to predict properties about a novel entity based on a known relation between that entity and some entity in $E$. Given some representative relation $r \in R$ that connects an entity $h \in E$ to a novel entity $e_n \notin E$, we can make predictions about its properties ($e_n$, $r_n$, $t_n$) using the 2-hop link ($h$, $r$, $r_n$, $?t$), where $?t \in E$. This generalization operates under the assumption that the relation connecting $h$ and $e_n$ is semantically meaningful such that we can make useful predictions about $e_n$ based on knowledge about $h$. The validity of this assumption may vary for different types of relations.


\section{Methods}

\mymethod{} is a case-based reasoning approach which aims to make predictions about new events based on background knowledge about similar cause-effect events. While our methods can be generalized to make predictions about new entities in other domains (conceptually replacing the ``cause event'' and ``effect event'' with different types of entities and connecting them by different relations as mentioned in the previous section), we describe our approach from the perspective of event prediction in this section.

At a high level, case-based reasoning aims to solve a new \textit{problem} based on experience and knowledge about similar \textit{cases} in the system's case base. In our task, the target problem refers to a new cause-effect event pair query for which we wish to make predictions, and a case refers to an event pair present in the KG. This approach can then be broken down as (1) retrieve cases from the KG which are similar to the new problem query, (2) identify and score paths through the KG that can be used to predict properties of the effect events in each case, and (3) reuse the learned paths to make predictions for our new problem.

\subsection{Case Retrieval} 

We define a case $s$ as a triple ($c_s$, $r$, $e_s$) connecting the cause $c_s$ to the effect $e_s$ by the causal relation $r$, where $(c_s, r, e_s) \in T$. In our running example KG, there are three example cases of cause and effect events connected by the \texttt{hasEffect} relation.

The first step of \mymethod{} is to retrieve cases from the KG that are similar to our problem, which we break down into two subtasks. First, we compute a similarity measure among entities in $E$ based on their subclass hierarchy and outgoing connections in the KG. Second, we compute the similarity of cases to the query $(c, r, e)$.

\subsubsection{Entity Similarity} 
To compute entity similarity, we apply a relatively simple similarity metric which can be computed based on count statistics and vector multiplication. Our goal in this step is to acquire a very rough sense of \textbf{similarity between individual entities}, which will be utilized in our subsequent steps.

For each entity $h_i \in E$, we form a vector representation $v_{h_i} = [a_1, a_2, ..., a_{N_h}]$, where $N_h$ is the total number of entities in $E$. Each value $a_j=1$ if there is a triple $(h_i, r, t_j) \in T$ for any relation $r$ or if $h_i$ is a subclass of $a_j$, and otherwise $a_j=0$. 

Intuitively, this vector allows us to capture the idea that similar entities are likely to be connected to the same entities. 

\paragraph{Example.}
The Kanto Earthquake and Tohoku Earthquake both have an outgoing connection to Japan, which might indicate that they are more similar than other earthquake events that occurred in different countries. Additionally, while these two earthquakes do not have any other outgoing connections that are identical, the \texttt{subclassOf} relation between Megathrust Earthquake and Earthquake indicates that both the Kanto and Tohoku Earthquakes are a type of Earthquake.
\qed

We also weight the importance of each entity in this vector based on how frequently it occurs as an outgoing neighbor or superclass. Our intuition here is to ensure that sharing outgoing connections to a less common entity is more meaningful than sharing connections to a common one. We compute this weight as an inverse document frequency (IDF) measure, $IDF(h) = \log(N_h / count(h))$, where $count(h)$ is the number of times $h$ has an incoming edge or is a superclass of an entity, and form the vector containing the IDF weighting of all entities as $v_{IDF} = [IDF(h_1), ..., IDF(h_{N_h})]$. Additionally, we apply normalization to $v_{IDF}$.

Lastly, we apply the IDF weighting to each vector by elementwise multiplication, and compute the similarity between two entities based on a weighted Jaccard similarity between those vectors. The entity similarity $ES$ is then computed as:

\begin{equation}
    wJaccard(x, y) = \frac{x \cdot  y}{\sum_{i=1}^{N_h} x_i + \sum_{i=1}^{N_h} y_i - x \cdot  y }
\end{equation}
\begin{equation}\label{eq:ent_sim}
    ES(h_i, h_j) = wJaccard(v_{h_i}\odot v_{IDF}, v_{h_j}\odot v_{IDF})
\end{equation}

\paragraph{Example.} The entity that is the most similar to the Tohoku Earthquake would be the Indian Ocean Earthquake, since both of them have outgoing connections to Megathrust Earthquake, Earthquake (through the subclass hierarchy), and Tsunami. While an entity like the Kanto Earthquake shares the outgoing connection to Japan, this link has a lower weighting since Japan has more connections in the KG compared to Megathrust Earthquake and Earthquake.
\qed

\subsubsection{Case Head Similarity} Next, we begin to determine the similarity of a case $(c_s, r, e_s)$ to our query $(c,r,e)$.

Our goal for case head similarity is to \textbf{determine how similar the case's cause is to \textit{c}.} Rather than directly using entity similarity from Equation \ref{eq:ent_sim}, we instead compute similarity based on the set of $c$'s outgoing triples, denoted as $(c, r_c, t_c) \in Out_{c}$. 

For each triple in $Out_{c}$, we define the importance of that triple based on (1) the probability that any triple in $T$ containing the tail entity $t_{c}$ also contains the relation $r_{c}$, denoted $P(r_{c}|t_{c})$, and (2) the probability that the tail entity $t_{c}$ occurs in any triple in $T$, denoted $P(t_{c})$. We posit that triples containing uncommon relations or that lead to uncommon entities should be considered more important.

The importance $I$ of a triple is then computed as:

\begin{equation}\label{eq:trip_importance}
    I(c, r_{c}, t_{c}) = \log \left( \frac{P(r_c | t_c)}{P(t_c)} \right)
\end{equation}

\paragraph{Example.}
The importance of ($c$, instanceOf, Megathrust Earthquake) will be greater than the importance of ($c$, country, Japan). The $P(r_c|t_c)$ terms for both triples will be equal to 1 in our KG snippet, while the $P(\text{Japan})$ is greater than $P(\text{Megathrust Earthquake})$.
\qed

After computing the importance of each triple in $Out_c$, we also normalize the values (denoted $nI(c, r_c, t_c)$) by dividing the importance of each triple by the sum of all importance values. 

Case head similarity of $c_s$ to the cause $c$ is then calculated as the weighted sum of similarities between each outgoing triple from $c_s$ (denoted $cOut_{s}$) and $Out_c$, using the most similar triple from each set for each unique relation. Denoting $ROut_c$ as the set of relations that occur in $Out_c$, we compute the case head similarity $CS_h$ as:

\begin{equation}\label{eq:head_sim}
    CS_h(c, c_s) = \sum_{r_c \in ROut_c} 
    \max_{\substack{(c_s,r_c,t_s) \in cOut_s \\ (c,r_c,t_c) \in Out_c}}( nI(c,r_c,t_c) * ES(t_c, t_s))
\end{equation}

\paragraph{Example.}
Tohoku Earthquake has the greatest $CS_h$ to our cause $c$, given that both the \texttt{instanceOf} and \texttt{country} have an exact match. Kanto Earthquake has the second highest $CS_h$, because it has an exact match with $c$ for the \texttt{country} relation leading to Japan and in its \texttt{instanceOf} relation, the entity similarity $ES$ between Earthquake and Megathrust earthquake is high. 
\qed

\subsubsection{Case Tail Similarity}
Besides judging how similar a case's cause event is to $c$, we also want some notion of \textbf{how similar the case's effect is to the effect \textit{e}}. Since $e$ is unseen in the KG and we do not actually know what the tail entities are for triples in $Out_e$, we determine similarity based on the types of outgoing relations from each entity. This similarity metric closely resembles those of works such as \cite{akbcCBR,ProbCBR}, which form a one-hot vector for each outgoing relation of entities to determine similarity.

We compute case tail similarity, $CS_t$, as a simple Jaccard similarity between the prediction relations, denoted $ROut_e$, and the outgoing relations for the case's effect, denoted $ROut_{es}$.

\begin{equation}
    CS_t(e_s) = \frac{|ROut_e \cap ROut_{es}|}{|ROut_e \cup ROut_{es}|}
\end{equation}

Lastly, we compute the coverage of the outgoing relations for the case tail, denoted $CC_t$. While $CS_t$ identifies case effects with similar relations to our effect event $e$, it also penalizes effects with a large number of relations. This second measure aims to ensure that we can identify cases whose tail entity can be used to make predictions about $e$ without applying such a penalty -- this measure might result in some entities that are less ``similar'' but more capable of performing reasoning through the KG.

\begin{equation}
    CC_t(e_s) = \frac{|ROut_e \cap ROut_{es}|}{|ROut_e|}
\end{equation}

\subsubsection{Case Selection}

Finally, we \textbf{select a number of similar cases to retrieve from the KG} which we will use for our case-based reasoning. After identifying the initial set of candidate cases, we score the similarity of the case to our query using Equation \ref{eq:case_sim_score} below. The cases are then sorted and we select the top $N_h$ cases. When selecting these top $N_h$ cases, we only select cases with a unique cause entity $c_s$ -- if another case with the same cause ranks among the top $N_h$ cases, it is disregarded. We follow this procedure to ensure a level of diversity in the cases over which we perform our reasoning.

\begin{equation}\label{eq:case_sim_score}
    CaseScore(c_s, r, e_s) = CS_h(c, c_s) * CS_t(e_s)
\end{equation}

We select an additional $N_t$ cases, where $N_t < N_h$, in which we prioritize the coverage of outgoing edges from the case's effect $e_s$. The motivation of this second set of cases is to ensure that we select some cases in which the case's effect contains as many relations in $ROut_e$ as possible. Therefore, we formulate the scoring equation to give $CC_t$ a greater influence than $CS_h$. We follow the same ranking and selection procedure as for the first $N_h$ cases, using Equation \ref{eq:case_cov_score} to rank the cases.

\begin{equation}\label{eq:case_cov_score}
    CaseScore_{cov}(c_s, r, e_s) = (1+CS_h(c, c_s)) * CC_t(e_s)
\end{equation}

We denote the set of cases selected through this procedure as $S$, containing $N_s=N_h+N_t$ total cases.

\paragraph{Example.}
In our KG snippet, we find that all of the cases' effect events have the same properties as those we wish to predict for $e$, and so our selection of the best cases will be judged by their head similarity $CS_h$ -- out of our three example cases, the Tohoku Earthquake case will have the highest $CaseScore$.
\qed

\subsection{Prediction Path Enumeration and Scoring}\label{sec:path_es}

We next use the retrieved cases to enumerate paths through the KG that can be used to make predictions about $e$.

\subsubsection{Path Enumeration}\label{sec:path_enum}

Our first step is to \textbf{enumerate a set of paths, using cases retrieved from the KG, which can be used to connect cause events to properties of their effect events}.
We define a path as a sequence of triples through the KG that can connect two entities. Given a start entity $x$ and end entity $y$, we can express a path of length $n$ connecting them as a sequence of triples, $p = [(x, r_1, e_1), ..., (e_{n-1}, r_n, y)]$. We define the \textit{relation path} as the sequence of relations used in each triple of a path, denoted $rel(p)=(r_1, ..., r_n)$. 
We denote the list of entities that can be reached starting from entity $x$ following the relation path $rel(p)$ as $\mathbb{E}_{p,x}$. Note that $\mathbb{E}_{p,x}$ is not a set, and may contain repeated entries of an entity that can be reached by different paths through the KG.

For each case $(c_s, r, e_s)$ in our set of retrieved cases $S$ and relation $r_e \in ROut_e$, let $E_{r_e,e_s}$ denote the set of entities connected to $e_s$ by the outgoing relation $r_e$, i.e., $(e_s, r_e, t) \in T$ for all $t \in E_{r_e,e_s}$. We then randomly sample up to $N_p$ unique paths of length $\leq 3$ connecting $c_s$ to any entity in $E_{r_e,e_s}$.

\begin{figure}[htbp]
    \centering
    \includegraphics[width=0.97\linewidth]{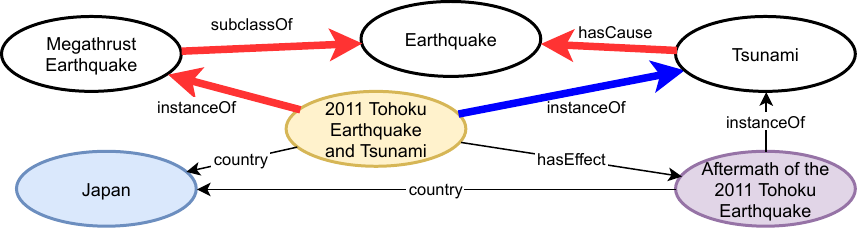}
    \caption{Example paths for the Tohoku Earthquake case.}
    \label{fig:pathsExamples}
\end{figure}

Here, we restrict the paths that are sampled such that (1) the first relation of $rel(p)$ must be an outgoing relation of our cause event $c$, and (2) the path does not traverse through the case's effect event $e_s$. Our first restriction ensures that we identify paths that are relevant to our query event -- since our goal is to follow relation paths starting from the event $c$, if the first relation is not an outgoing relation of $c$ it will not be possible to utilize the path. Also note that the triple $(c, r, e)$ is not present in the KG, so a path starting with $r$ would only be valid if $c$ has some other outgoing triple with the relation $r$. Similarly, our second restriction aims to prevent us from sampling paths that traverse through the causal relation connecting the cause and effect event. Again, since $e$ is unseen in the KG, such a path would not provide us with any useful information to apply to our event query.

\paragraph{Example.}
Figure \ref{fig:pathsExamples} shows two example paths, highlighted in red (light) and blue (dark), connecting the Tohoku Earthquake event to its effect's \texttt{instanceOf} property.
$rel(p_{red})$ consists of the three relations [instanceOf, subclassOf, hasCause$^{-1}$] while $rel(p_{blue})$ is the single relation [instanceOf]. We do not sample paths that traverse through the Aftermath event.
\qed

The path sampling steps are repeated for each case in $S$, and the set of all paths sampled from each case, connecting the case's cause to its effect's $r_e$ properties, is denoted as $\mathcal{P}_{S, r_e}$.

\subsubsection{Path Scoring} Next, we \textbf{compute confidence scores for each unique relation path} present in $\mathcal{P}_{S,r_{t_q}}$. This confidence score will correspond to the notion of how confident we are that a given path leads to the correct entity prediction.

We base our scoring on a simple precision measure, aiming to score how well a given relation path leads to the correct entities for the target relation. Additionally, following from the path confidence measure implemented in \cite{AnyBURL}, we add a smoothing constant $\epsilon$ (set to $\epsilon=5$ in our experiments) to the denominator when calculating the precision -- this allows for relation paths with the same precision and a greater number of samples to have a higher score than relation paths with fewer samples. The relation path score for a target relation to predict, $r_e$, and relation path $rel(p)$ is given as:

\begin{equation}\label{eq:path_score_forward}
    PScore(r_e, rel(p)) = \frac{
    \sum_{(c_s, r, e_s) \in S} \sum_{t' \in \mathbb{E}_{p, c_s}} \mathds{1}[t' \in E_{r_e, e_s}]
    }
    {\epsilon + 
    \sum_{(c_s, r, e_s) \in S} |\mathbb{E}_{p, c_s}| }
\end{equation}

where $\mathds{1}[t' \in E_{r_e, e_s}]=1$ if $t' \in E_{r_e, e_s}$ is true and 0 otherwise.

The $PScore$ for a particular $rel(p)$ may be influenced by sampling multiple different paths that have the same relation path. This allows us to implicitly add a weighting to each relation path based on how frequently they occur, which corresponds to how likely they are to be randomly sampled.

\paragraph{Example.}
To make a prediction about the effect's \texttt{country} in our three cases, we can sample the relation path $rel(p)$ = [country]. For all three of our cases, following the relation path [country] from the cause leads to the correct country property of its effect. The $PScore$ of this path would then be $3/(\epsilon+3)$. On the other hand, for a relation path like $rel(p)$ = [instanceOf] from our example in Figure \ref{fig:pathsExamples}, which aims to predict the \texttt{instanceOf} relation of the effect, the correct entity is only reached in 1 out of 2 results for the Tohoku Earthquake, and 0 out of 3 total results for the other two cases. The $PScore$ for this path would then be $1/(\epsilon+5)$.
\qed

Based on these path scores, produced for the set of distinct relation paths in $\mathcal{P}_{S, r_e}$, we select $N_p$ relation paths with the highest scores, denoted $\mathcal{P}_{r_e}$, to make predictions for $(e, r_e, ?z)$.

\subsection{Applying Prediction Paths}\label{sec:fwd_score}

Given our set of relation paths $\mathcal{P}_{r_e}$, selected and scored using our set of retrieved cases, we can now \textbf{apply these paths to our cause event \textit{c} to make predictions about property $r_e$ of the effect event}. We perform these predictions by following each relation path, starting from our cause $c$, and using the $PScore$ for each path to produce a total confidence score for each predicted entity. 

Formally, for our query event $(c, r, e)$ and a single property for which we want to predict $(e, r_e, ?z)$, we score each candidate prediction entity $z$ as follows:

\begin{equation}\label{eq:ent_pred_score_fwd}
    EScore(c, r, r_e, z) = \sum_{
    \substack{
    rel(p) \in \mathcal{P}_{r_e} \\ z' \in \mathbb{E}_{p, c}
    }
    } PScore(r_e, rel(p))[z'=z]
\end{equation}

In Equation \ref{eq:ent_pred_score_fwd}, we choose to calculate the score for a prediction $z$ as a summation of path $PScore$s so that the score is increased for (1) entities that are frequently reached by a given relation path, and (2) entities that are reached by a variety of different relation paths.

For the target relation $r_e$ we compute all $EScore$ values for all entities for which a path $rel(p) \in \mathcal{P}_{r_e}$ exists between it and $c$, denoted $E_{\mathcal{P}_{r_e}, c}$. We use these scores to sort and rank our predictions for the given 2-hop link. To make predictions for all properties $Out_e$ of that we wish to predict, we repeat the procedures in Section \ref{sec:path_es} for each $r_e \in ROut_e$ and the aforementioned scoring procedure.

\paragraph{Example.}
Let us consider applying paths to make predictions for $e$'s \texttt{instanceOf} relation. There are only two valid paths that may be sampled to make this prediction, $rel(p_1)$ = [instanceOf] and $rel(p_2)$ = [instanceOf, subclassOf, hasCause$^{-1}$], both coming from the Tohoku Earthquake case. Setting $\epsilon=0$ for simplicity, the $PScore$ for these two path are $PScore(p_1) = 1/5$ and $PScore(p_2) = 1$. Applying these two paths to our cause event $c$ results in predicting $e$ to be an instance of $Megathrust Earthquake$ with a score of $1/5$ and $Tsunami$ with a score of 1. The best prediction in our minimal running example is that the new effect event will be a tsunami.
\qed

Algorithm \ref{alg:mymethod} provides a high-level overview of our model's inputs and the steps that are performed to produce predictions for each property of the unseen effect event up to this point.

\begin{algorithm}\caption{\mymethod{} Event Prediction Overview}\label{alg:mymethod}
\begin{algorithmic}[1]
\Statex \textbf{Inputs:} $(c,r,e), ROut_e, N_h, N_t, N_p$
\Statex $(c,r,e) \gets$ the cause $c$, causal relation $r$, and unseen effect $e$
\Statex $ROut_e \gets$ the set of relations about the effect to predict
\Statex $N_h \gets$ number of cases to retrieve using $CaseScore$
\Statex $N_t \gets$ number of cases to retrieve using $CaseScore_{cov}$
\Statex $N_p \gets$ number of paths to sample
\State Retrieve cases from the KG
\For{$r_e \in ROut_e$} \Comment{Repeat for each prediction relation}
\State Sample $N_p$ paths from each case
\State Compute each path's score using $PScore$
\State Follow each path starting from $c$ to generate predictions
\State Score each prediction using $EScore$
\State Rank and report top predictions for $r_e$
\EndFor
\end{algorithmic}
\end{algorithm}

\subsection{Prediction Score Refinement}\label{sec:backward_score}

Optionally, after making predictions for all of our event prediction relations $ROut_e$, we introduce an additional step to \textbf{refine our prediction rankings} produced by Equation \ref{eq:ent_pred_score_fwd}. Given our task setup, in which we perform a prediction based on some properties about a cause event, our refinement step aims to apply our prediction methods in the opposite direction to \textbf{make ``predictions'' about the \textit{cause} event's properties starting from the \textit{effect} event}.

The key intuition behind this step is that if we have chosen the correct entity $z$ for the prediction triple $(e, e_r, ?z)$, we would expect that applying our prediction methods starting from the event $e$ to predict properties of $c$ should yield good results. Furthermore, because we know the true properties of $c$, we can refine the score of the prediction $(c, r, r_e, ?z)$ based on how accurately it can be used to perform this reverse prediction. 

\begin{figure}[htbp]
    \centering
    \includegraphics[width=\linewidth]{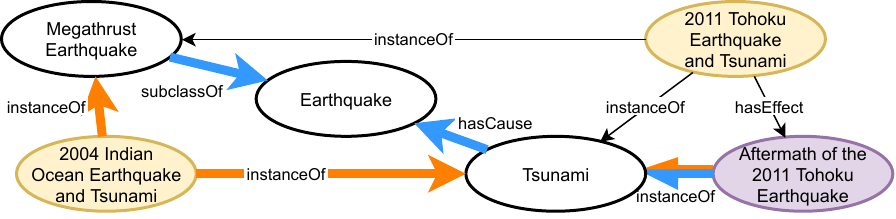}
    \caption{Example paths connecting the effect event (Aftermath of...) to the Tohoku Earthquake's \texttt{instanceOf} property.}
    \label{fig:motivate_refinement}
\end{figure}

\paragraph{Example.}
A visualization of our refinement step is shown in Figure \ref{fig:motivate_refinement}, for the case of the Tohoku Earthquake event. Two paths connecting the effect to the cause's \texttt{instanceOf} property are highlighted in orange (light) and blue (dark).
Similar to sampling paths from this case to make predictions about the effect, we now aim to sample paths to make predictions about the cause. 
\qed

To perform our refinement step we reuse our set of previously retrieved cases $S$ and proceed with producing refined scores for one prediction relation $r_e$ at a time. Following through the path enumeration and scoring methods of Section \ref{sec:path_es}, we now produce paths for each unique outgoing relation of the \textit{cause} event. Let $ROut_c$ denote the set of outgoing relations from our cause event $c$. For each $r_c \in ROut_c$, we sample a set of paths $\mathcal{P}_{S^{-1}, r_c}$ indicating paths connecting $e_s$ to any entity in $E_{r_c, c_s}$ for each case $(c_s, r, e_s) \in S$. We apply similar restrictions on our path sampling as in Section \ref{sec:path_enum}, sampling up to $N_p$ paths, limiting paths' length to 3 relations, and only selecting paths whose first relation is in the set of prediction relations $ROut_e$.

For each relation path in $rel(p) \in \mathcal{P}_{S^{-1}, r_c}$ , we produce a path score $PScoreR$, as:

\begin{equation}\label{eq:path_score_backward}
PScoreR(r_c, rel(p)) = 
    \frac{
    \sum_{(c_s, r, e_s) \in S} \sum_{t' \in \mathbb{E}_{p, e_s}} \mathds{1}[t' \in E_{r_c, c_s}]
    }
    {\epsilon + 
    \sum_{(c_s, r, e_s) \in S} |\mathbb{E}_{p, e_s}| }
\end{equation}

We note that $PScoreR$ is identical to $PScore$ from Equation \ref{eq:path_score_forward}, except that the starting points of paths are switched to reflect our swap to following paths from the effect to the cause event.

For each predicted entity $z \in E_{\mathcal{P}_{r_e}, c}$ which we produced using the methods up to Section \ref{sec:fwd_score}, we produce a sequence refinement score, $RS(z, r_e, (c, r_c, t_c))$, corresponding to how accurately paths that contain the entity $z$ can predict outgoing triples of $c$, where $(c, r_c, t_c) \in Out_c$. Additionally, we temporarily treat the prediction triple $(e, r_e, z)$ as being present in the KG so that we can follow paths starting from our unseen effect event $e$. Since we only temporarily add this single triple, only paths that traverse through the prediction entity $(e, r_{e}, z)$ will be considered in calculating $RS$.

\begin{equation}\label{eq:refined_score}
    RS(z, r_e, (c, r_c, t_c)) = \sum_{
    \substack{
    rel(p) \in \mathcal{P}_{S^{-1}, r_c} \\ t' \in \mathbb{E}_{p, e}
    }
    } \frac{PScoreR(r_c, rel(p))[t'=t_c]}{|\mathbb{E}_{p, e}|}
\end{equation}

We normalize these values by dividing each $RS$ by the maximum $RS$ score produced for each triple -- i.e., for each triple in $Out_c$ and entity $z \in E_{\mathcal{P}_{r_e}, c}$, the normalized refinement score $nRS$ is calculated:

\begin{equation}
    nRS(z, r_e, (c, r_c, t_c)) = \frac{RS(z, r_e, (c, r_c, t_c))}{\max_{z_i \in E_{\mathcal{P}_{r_e}, c}}(RS(z_i, r_e, (c, r_c, t_c))}
\end{equation}

$nRS$ now provides us with a score in the range of [0,1] which indicates how well a given entity $z$ can ``predict'' a particular cause event, where the $nRS=1$ for the entity $z$ that produces the highest $RS$ score for a particular triple $(c, r_c, t_c)$. We then compute our final refinement score as:

\begin{equation}\label{eq:ent_pred_score_bkwd}
\begin{aligned}
    nRS_{max}&(z, r_e) = \max_{(c, r_c, t_c) \in Out_{c}}(nRS(z, r_e, (c, r_c, t_c))) \\
    nRS_{avg}&(z, r_e) = \frac{1}{|Out_c|}*\sum_{(c, r_c, t_c) \in Out_c}(nRS(z, r_e, (c, r_c, t_c))) \\
    ReScore&(h_q, r_q, r_e, z) = \\ 
    &EScore(h_q, r_q, r_e, z) * ( nRS_{max}(z, r_e) + nRS_{avg}(z, r_e))
\end{aligned}
\end{equation}

In Equation \ref{eq:ent_pred_score_bkwd}, high values of $nRS_{max}$ provides us with evidence that $z$ can be used to predict \textit{some} triple in $Out_c$ well, while high values of $nRS_{avg}$ indicate that $z$ can predict many triples in $Out_c$ well. We combine these scores so that we can reward prediction entities that can predict all of the cause's properties while not over-penalizing entities that still are able to reach some of the cause's properties with high accuracy. 

\paragraph{Example.}
Our initial predictions for $e$'s \texttt{instanceOf} relation were Tsunami and Megathrust Earthquake. To refine the scores of these two predictions, our method determines how well we can reach $(c, \text{country}, \text{Japan})$ and ($c$, instanceOf, Megathrust Earthquake). Using paths such as the two shown in Figure \ref{fig:motivate_refinement}, we find that the Tsunami prediction can more accurately reach $c$'s \texttt{instanceOf} property compared to the Megathrust Earthquake prediction. This leads to Tsunami receiving a higher $nRS_{max}(z, instanceOf)$ score, which subsequently leads to a higher $ReScore$ value.
\qed

After producing the refined scores for each entity, we re-rank them to produce our refined predictions. For each prediction relation $r_e$, the above process is repeated. 

\section{Experiments}

To evaluate our work, we perform a modified evaluation over a commonly used benchmark dataset, FB15k-237~\cite{fb15k237}, as well as a novel dataset curated from Wikidata. Model performance is measured in terms of the predicted tail entity for a given 2-hop link. 

\subsection{Datasets}

Given our problem setup as a 2-hop LP task for properties of unseen entities, we split our test data based on entities rather than individual triples.
We split both datasets into training triples, validation connections, validation triples, test connections, and test triples. We refer to any entity contained in the training triples as part of the training set, while head entities of triples in the test and validation triples are part of the test and validation sets, respectively.
The test connections indicate relations from an entity in the training set to an entity in the test set. The test triples then indicate outgoing triples from an entity in the test set to one in the training set. 

\subsubsection{FB15k-237} For the FB15k-237 dataset, we iteratively select random entities to place in the test set based on the following conditions: for a candidate test entity $x$, (1) there must be no triple connecting $x$ to an entity in the test set, (2) $x$ must have at least one incoming and outgoing triple, and (3) any entity connected to $x$ must have at least one other triple connecting it to a different entity in the training set. 
These conditions ensure that test triples can plausibly be predicted by models and that all training entities have training data available.
After selecting entities to place in the test set, we randomly select half of those for our validation set. 

From the 14,541 entities in FB15k-237, we select 500 entities each for the test and validation sets. 
The connecting and test triples together combine to form a total of 170,000 2-hop links over which we perform evaluation. 

\subsubsection{Wikidata Events} 
To curate our Wikidata causal event dataset, we use an approach similar to~\cite{Hassanzadeh21a} to identify 307 event-related classes based on links between entries in Wikidata and news articles in Wikinews.
Following Wikidata's guidelines on modeling causes, we select 6 causal relations\footnote{\url{https://www.wikidata.org/wiki/Wikidata:List_of_properties/causality}} which we then use to query Wikidata for pairs of event entities that are connected by a causal relation.

This query yielded a set of \textbf{1,953} pairs of events, encompassing 157 unique event classes. These triples connect \textbf{284} unique ``cause'' events to \textbf{311} unique ``effect'' events. We then collect the 3-hop neighborhood of outgoing connections surrounding each of these event entities, removing all literals (e.g., strings, integers) and any entity for which only one triple existed in the dataset. 
Our final dataset consists of \textbf{758,857} triples for \textbf{123,166} entities.

From our set of cause-effect event pairs, we randomly select 100 effect events to serve as the test set. After filtering to remove test events related to each other, we end up with a test set consisting of 89 unique effect events, with 104 connections to causal events to test over, corresponding to a total of 1,365 2-hop links to evaluate.

\subsection{Baseline Models}

To compare \mymethod{}'s performance against baselines, we modify each model's scoring function to incorporate the 2-hop link. For instance, for TransE \cite{transe}, which aims to learn embeddings for $(h,r,t)$ by optimizing for $h+r = t$, we can score a 2-hop prediction as $h+r_1 = t_1$, $t_1+r_2 = t_2$ $\xrightarrow{}$ $h+r_1+r_2 = t_2$.
We follow a similar procedure for each of our baseline models which rely on learning embeddings.
This modified scoring is only used for testing, and model training is performed normally using the training triples.

For our baselines, we choose three embedding-based models which have seen widespread use in recent years -- TransE \cite{transe}, ComplEx \cite{trouillon2016complex}, and RotatE \cite{sun2019rotate}. For each model, we test embedding sizes of \{100, 200, 300\} dimensions and train the model for 100 iterations\footnote{We select these hyperparameters to explore based on prior experience and related work, and consider further fine-tuning \cite{Teach2020YouCT} to be beyond the scope of this research.} using self-adversarial negative sampling \cite{sun2019rotate}.

We also compare against NoGE \cite{NoGE}, a graph neural network (GNN) model which models both entities and relations as nodes in the graph. NoGE was developed and evaluated on a Wikidata-based dataset, CoDEx \cite{codexDataset}, which we believe makes it a strong candidate for showing good performance on our own causal event dataset. We train  NoGE for 100 iterations using default configurations, 
and test using embedding sizes of \{64, 128, 256\}. 

Lastly, we compare against two rule-based baselines: ProbCBR \cite{ProbCBR} and AnyBURL \cite{AnyBURL}. ProbCBR is a similar model to ours, which leverages clustering of entities to better estimate scores of reasoning paths through the KG. We perform experiments with ProbCBR using parameter settings of retrieving \{5, 10, 20\} cases and sampling \{60, 80, 100\} paths. AnyBURL on the other hand is a rule learning model which can efficiently sample the KG to learn and generalize logical rules. Following from the original publication, we train AnyBURL for 1,000 seconds with default parameters. 

For our own models, we report our results for three variations of our entity scoring. \mymethod{}$_{base}$ is our basic case-based reasoning approach, following the path sampling and scoring methods up to Section \ref{sec:fwd_score}. \mymethod{}$_{re}$ denotes the use of our score refinement method, from Section \ref{sec:backward_score}. Lastly, \mymethod{}$_{re+base}$ indicates the addition of the scores produced by the base and refinement methods. We evaluate our methods by retrieving $N_h=\{5, 10, 20\}$ cases using $CaseScore$, $N_t=\{1,3,5\}$ cases using $CaseScore_{cov}$, and $N_p=\{60, 80, 100\}$ sample paths.

\subsection{Results and Discussion}

We apply each model to rank predictions for each tail entity in the test set of 2-hop links and report the mean reciprocal rank (MRR) and Hits@K metrics for each model. Experimental results 
are shown in Table \ref{tab:2hop_exp_combined}; the best performing model for each dataset is highlighted in bold and the second best is underlined (for \mymethod{}'s results we only highlight the best performance of any one variation). 


\begin{table}[htb]
    \centering
    \caption{Results for our 2-hop LP experiments.}
    \label{tab:2hop_exp_combined}
    \begin{tabular}{|l|c|c|c|}
    \hline
    \multicolumn{4}{|c|}{Dataset: FB15k-237} \\
    \hline
    Model & MRR & Hits@1 & Hits@10 \\
    \hline
    TransE & 0.355 & 0.253 & \underline{0.553} \\
    ComplEx & 0.151 & 0.101 & 0.249 \\
    RotatE & 0.290 & 0.191 & 0.477 \\
    NoGE & 0.324 & 0.238 & 0.491 \\
    AnyBURL & \textbf{0.381} & \textbf{0.289} & \textbf{0.560} \\
    ProbCBR & 0.289 & 0.204 & 0.452 \\
    \hline
    \mymethod{}$_{base}$ & 0.364 & 0.273 & 0.537 \\ 
    \mymethod{}$_{re}$ & 0.349 & 0.256 & 0.524 \\ 
    \mymethod{}$_{re+base}$ & \underline{0.368} & \underline{0.277} & 0.543 \\ 
    \hline
    \hline
    \multicolumn{4}{|c|}{Dataset: Wikidata Causal Events} \\
    \hline
    Model & MRR & Hits@1 & Hits@10 \\
    \hline
    TransE & 0.139 & 0.096 & 0.207 \\
    ComplEx & 0.030 & 0.024 & 0.050 \\
    RotatE & 0.097 & 0.081 & 0.125 \\
    NoGE & \underline{0.149} & 0.118 & \underline{0.210} \\
    AnyBURL & \underline{0.149} & \underline{0.121} & 0.202 \\
    ProbCBR & 0.122 & 0.107 & 0.154 \\
    \hline
    \mymethod{}$_{base}$ & 0.156 & 0.118 & 0.222 \\
    \mymethod{}$_{re}$ & 0.158 & \textbf{0.130} & 0.212 \\
    \mymethod{}$_{re+base}$ & \textbf{0.159} & 0.125 & \textbf{0.227} \\
    \hline
    \end{tabular}
\end{table}

In the causal event data, we find that variations of \mymethod{} show superior performance over baselines, while for FB15k-237 \mymethod{} shows second-best performance. In particular, we observe that applying our refinement step \mymethod{}$_{re}$ leads to the best Hits@1 performance for the event dataset, while \mymethod{}$_{re+base}$ shows the highest MRR and Hits@10. 
\mymethod{}$_{re}$'s lower Hits@10 may be attributed to situations where refinement fails to sample paths connecting a prediction entity to the input cause's properties (i.e., leading to $nRS_{max}(h, r_e)=0$, which subsequently makes its $ReScore$ equal 0).

Our results suggest that \mymethod{} shows strong performance for the task of event prediction, and more generally for the task of 2-hop LP for properties about unseen entities. Our model does not require any training, which further bolsters its applicability to this task in the open-world setting, where we might see frequent changes to the underlying KG. In all of our experiments, the performance of \mymethod{} is reflective of making predictions for the effect event if the cause event's properties were newly added to the KG. In contrast, all baselines except ProbCBR perform training while including the cause events in the training set. To compare the impact of this fact, for our AnyBURL baseline, if we remove all learned rules that explicitly refer to the cause entity when performing predictions, MRR decreases to 0.363 and 0.127 for the FB15k-237 and causal event datasets, respectively -- under this condition, \mymethod{} now outperforms AnyBURL on the FB15k-237 dataset.

As an example of applying \mymethod{} to ongoing events, we performed a prediction for the effect of a Protest event in Iran. Similar cases retrieved from the KG included the Bahraini Protests of 2011, the Iranian Revolution, and the 2019-21 Chilean Protests. The top \texttt{instanceOf} predictions for the effect were Resignation, Demonstration, and Civil Resistance, while the top \texttt{country} predictions were Iran, Iraq, and Azerbaijan. We find that our method shows promise in terms of factors such as predicting the possibility of events in Iraq and Azerbaijan, which both are geographically and politically intertwined with Iran, and are already dealing with consequences of the events in Iran. Our method also performs well in retrieving cases of past events that have similar causes and likely similar consequences.

On the other hand, we also observe some situations in which \mymethod{} struggles. One failure pattern was when no particularly similar event pairs were present in the KG -- this was not uncommon due to Wikidata's variable coverage and level of detail. Another noticeable issue was the overabundance of COVID-19 related events in Wikidata, which frequently were retrieved by \mymethod{} due to matching the target \texttt{country} of a query event.

\section{Conclusion}

We introduce \mymethod{}, a case-based reasoning model developed to perform event prediction between a causal event and its unseen effect event. Framing event prediction as a 2-hop link prediction task, \mymethod{} retrieves cases of cause-effect events from the KG to sample and apply paths through the KG that connect causes to their respective effect's properties. A novel refinement step helps improve the accuracy of predictions. 
\mymethod{} does not require training, making it well suited to our intended application to KGs under the open-world assumption. We evaluate the effectiveness of \mymethod{} over the FB15k-237 dataset as well as a newly curated dataset of causal events from Wikidata, 
showing strong performance compared to baselines consisting of three embedding-based models, a GNN model, and two rule-based models which use similar scoring and sampling methods as ours.
Future work should continue to explore the practically important application of event prediction using KGs, as well as make further advances in reasoning techniques over constantly evolving KGs such as Wikidata.

\begin{acks}
This work was supported by the \href{http://airc.rpi.edu}{Rensselaer-IBM AI Research Collaboration}, part of the \href{http://ibm.biz/AIHorizons}{IBM AI Horizons Network}.
\end{acks}

\balance

\bibliographystyle{ACM-Reference-Format}
\bibliography{sample-base}


\end{document}